\documentclass[pdflatex,sn-mathphys-num]{sn-jnl}

\usepackage{graphicx}%
\usepackage{multirow}%
\usepackage{amsmath,amssymb,amsfonts}%
\usepackage{amsthm}%
\usepackage{mathrsfs}%
\usepackage[title]{appendix}%
\usepackage{xcolor}%
\usepackage{textcomp}%
\usepackage{manyfoot}%
\usepackage[normalem]{ulem}
\usepackage{booktabs}%
\usepackage{algorithm}%
\usepackage{algorithmicx}%
\usepackage{algpseudocode}%
\usepackage{listings}%
\usepackage{adjustbox}

\theoremstyle{thmstyleone}%
\theoremstyle{thmstyletwo}%

\theoremstyle{thmstylethree}%

\begin{document}

\title[Article Title]{Exploring Remote Photoplethysmography for Neonatal Pain Detection from Facial Videos}


\author[1]{\fnm{Ashutosh} \sur{Dhamaniya}}\email{ms2204101005@iiti.ac.in}
\equalcont{These authors contributed equally to this work.}

\author*[1]{\fnm{Anup Kumar} \sur{Gupta}}\email{msrphd2105101002@iiti.ac.in}
\equalcont{These authors contributed equally to this work.}

\author[1]{\fnm{Trishna} \sur{Saikia}}\email{phd2201101014@iiti.ac.in}

\author[1]{\fnm{Puneet} \sur{Gupta}}\email{puneet@iiti.ac.in}

\affil[1]{\orgdiv{Department of Computer Science and Engineering}, \orgname{Indian Institute of Technology Indore}, \orgaddress{\postcode{452020}, \country{India}}}

\abstract{Unaddressed pain in neonates can lead to adverse effects, including delayed development and slower weight gain, emphasising the need for more objective and reliable pain assessment methods. Hence, automated methods using behavioural and physiological pain indicators have been developed to aid healthcare professionals in the Neonatal ICU. Traditional contact-based methods for physiological parameter estimation are unsuitable for long-term monitoring and increase the risk of spreading diseases like COVID-19. We introduce a novel approach using remote photoplethysmography (rPPG)  to estimate pulse signals in a non-contact manner and employ them for neonatal pain detection. The temporal signals acquired from regions-of-interest (ROIs) affected by skin deformations may exhibit lower quality and provide erroneous rPPG signals. Therefore, we incorporated a quality parameter to select the temporal signals obtained from ROIs that are least affected by skin deformations. Further, we employed signal-to-noise ratio as a fitness parameter to extract the rPPG signal corresponding to the clip that is least affected by noise. Experimental findings demonstrate that the rPPG signals provide useful information for neonatal pain detection, and signals extracted from the blue colour channel outperform those extracted from other colour channels. We also show that combining rPPG and audio features provides better results than individual modalities.}


\maketitle

\section{Introduction}\label{sec1}

Pain detection in neonates is a critical aspect of paediatric healthcare. Neonates, especially those in intensive care units, are often subject to procedures that can cause pain \cite{brahnam2023neonatal, zamzmi2017review, cheng2022artificial}. However, their inability to communicate verbally makes pain assessment challenging \cite{brahnam2023neonatal, zamzmi2016approach}. Unfortunately, unaddressed pain in neonates can lead to various short-term and long-term complications, including altered pain sensitivity, developmental delays, changes in the gray and white matter of the brain, as well as emotional and psychological issues later in life \cite{page2004there, mathews2011pain, brahnam2023neonatal}. The current approach to pain assessment relies on the observation of caregivers for specific physiological and behavioural pain indicators \cite{zamzmi2017review}. The pain assessment by caregivers is significantly influenced by various factors such as the cognitive bias of the observer, background, gender, identity, and culture \cite{zamzmi2017review, brahnam2023neonatal}. These factors may result in inconsistent pain assessment and treatment. Hence, an accurate and timely detection of pain is crucial for pain management, as well as the overall well-being of the neonate. However, considering the current constraints in healthcare staffing, evidenced by a substantial shortage of medical personnel globally \cite{meyer2023staffing, zamzmi2017review}, coupled with the overwhelming workload on existing staff, the development of automated pain detection systems has become imperative.

Existing automated pain assessment methods in neonates primarily rely on behavioural indicators, such as facial expressions, crying, and body movements \cite{brahnam2023neonatal, ritu2022facial, sun2022aue, lu2023video, salekin2019harnessing}. A few recent studies have included physiological indicators like heart rate (HR), heart rate variability (HRV), and breathing rate for pain assessment \cite{zamzmi2016approach, zamzmi2017automated, jiang2024personalized}. These physiological measures offer several advantages over behavioural indicators. While behavioural measures are reported to be powerful indicators of pain, they have limitations, especially in infants who may have a reduced ability to express pain behaviourally due to certain disorders or physical conditions, like post-surgery exhaustion \cite{zamzmi2016approach}. This underlines the necessity of incorporating physiological measures for a more comprehensive assessment of neonatal pain. Traditionally, monitoring of vital signs in the Neonatal Intensive Care Unit (NICU) has relied on adhesive electrodes and sensors attached to the skin of the neonate \cite{zamzmi2017review}. These sensors record vital signals, which are then converted to a format suitable for display and monitoring. However, this contact-based method can be intrusive and uncomfortable for neonates, potentially interfering with their natural state and responses \cite{lokendra2022and, gupta2020mombat, gupta2023radiant}.

In contrast to contact-based methods, remote photoplethysmography (rPPG) presents a non-contact alternative for capturing physiological data such as HR \cite{saikia2023hreadai, gupta2023radiant}, HRV \cite{li2019improvement}, respiratory rate (RR) \cite{alnaggar2023video}, and blood oxygen saturation (SpO2) \cite{agarwal2025shine}. It works on the principle that there is a subtle change in the skin colour of the neonates as blood pumps through it. Digital cameras can capture these subtle variations in light reflected by the skin, enabling HR monitoring without direct physical contact with the neonate \cite{huang2021neonatal}. This way, rPPG alleviates the physical discomfort associated with adhesive sensors and minimises the risk of infection, a crucial consideration in the delicate environment of the NICU \cite{malafaya2020domain}. However, the performance of the rPPG signal heavily relies on the colour channel, which is used as temporal signal to estimate the rPPG signal. Moreover, it is susceptible to various inevitable noises, such as eye blinking, frequent movements of the face and body, illumination variations, and skin deformations. These factors introduce significant skin colour fluctuations, resulting in erroneous HR estimation.

To the best of our knowledge, the applicability of rPPG in neonatal pain detection has not been studied. Also, we believe that despite several challenges in rPPG estimation, it is worth exploring rPPG to detect pain in neonates who cannot communicate their pain verbally due to its non-invasiveness and ability to provide continuous monitoring. Our contributions can be summed up as follows: (1) To the best of our knowledge, we are the first to explore the rPPG signal extracted from facial videos for neonatal pain classification. (2) In the literature, the green colour channel has been considered to contain the strongest photoplethysmographic information. However, our experimental findings indicate that the blue colour channel surpasses the performance of other colour channels for neonate pain detection. (3) We divided the input video into multiple clips to avoid the inevitable noises available in long-duration videos. In these clips, the temporal signals obtained from the region of interest (ROIs) affected by skin deformations may show lower quality due to significant colour variations in the skin. Therefore, we employed a quality parameter to select temporal signals least impacted by skin deformations. Moreover, we have utilised the signal-to-noise (SNR) ratio as a fitness parameter to evaluate the quality of the rPPG signal obtained from each clip. Among all the obtained rPPG signals, we selected the one least affected by the unavoidable noises. (4) Our experimental results demonstrate that rPPG provides useful information for neonatal pain detection. Moreover, we also show that the fusion of the rPPG signal with audio features performs better than the individual modalities.

\section{Related Work}

In recent years, advancements have been made in automating neonatal pain assessment, addressing the shortcomings of traditional methods. The automated methods rely either on behavioural, physiological, or multimodal features for pain detection. The methods based on behavioural features rely on pain indicators such as facial expression, facial action units, body movement, and crying sounds. The methods based on physiological features rely on vital signs such as HR, HRV, respiratory rate (RR), and oxygen saturation ($\text{SpO}_2$) for pain detection. Meanwhile, the multimodal approaches combine two or more behavioural or physiological features for neonatal pain detection. 

Initial studies focused on preprocessing face images through cropping, grayscale conversion, and converting them to 1D vectors. Techniques like Linear Discriminant Analysis (LDA) and Principal Component Analysis (PCA) were employed to reduce the dimensionality of these vectors, with the condensed features then classified using Support Vector Machines (SVM) \cite{brahnam2006svm, brahnam2006machine}. The study proposed in \cite{schiavenato2008neonatal} manually identified facial points based on the Neonatal Facial Coding System \cite{grunau1987pain}; using the distances between these points for pain prediction. Recent works have employed Gaussian of Local Descriptors (GOLD) and Bag-of-Features (BoF) for handcrafted feature extraction, utilising SVM classifiers for pain detection \cite{brahnam2023neonatal}. Similarly, in  \cite{ritu2022facial}, handcrafted features, such as facial geometric features and Local Binary Patterns (LBP), are fed to an SVM for pain detection. Moreover, combining deep features from Convolutional Neural Networks (CNN) with traditional handcrafted features like LBP and Histogram of Oriented Gradients (HOG) has shown significant improvements in pain classification \cite{celona2017neonatal}. Comparative studies have also demonstrated that deep learning (DL) architectures, particularly CNN variants, surpass traditional features such as LBP, HOG, and Gabor when applied to SVM \cite{yan2020fenp}. Further, pre-trained ResNets and two-stream architectures incorporating spatial and temporal information through cross-stream attention mechanisms have been explored for pain classification \cite{sun2022aue, lu2023video, salekin2020first}. Transfer learning from pre-trained CNNs has also been investigated for feature extraction in pain detection \cite{lu2018deep, zamzmi2018neonatal}.

Apart from facial features, body movement, and crying audio sounds have been identified as strong indicators of pain detection in neonates. Frameworks combining body movement and facial expressions using SVM, CNN, and Long Short-Term Memory (LSTM) networks have been proposed, as well as unsupervised fusion of facial and body skeleton sequences for pain intensity prediction \cite{salekin2019multi, zhu2023video, sun2019automatic}. Audio features such as spectrogram from crying sounds have been analysed using CNNs and SVMs for neonatal pain assessment \cite{salekin2019harnessing, ashwini2021deep}. Recent works integrate audio-visual features for a more comprehensive pain assessment \cite{salekin2021multimodal}.

In several studies, researchers have explored how physiological parameters are correlated to pain stimuli, and the findings indicate a strong association between pain and physiological parameters such as HR, HRV, and others \cite{raeside2011physiological, lindh1999heel, faye2010newborn, weissman2012heart}. Consequently,  physiological pain indicators are used in conjunction with behavioural indicators like facial expressions and body movement to assess neonatal pain. For instance, in the study \cite{zamzmi2016approach}, strain-based features from facial expressions and motion features are employed to obtain pain scores. On the other hand, statistical features from physiological parameters are fed to a Random Forest classifier for pain scores. Finally, majority voting is applied to scores based on behavioural and physiological pain indicators to assess the pain. This study was extended to include audio modality, utilising features like Linear Prediction Cepstral Coefficients and Mel-frequency Cepstral Coefficients (MFCC), enhancing the multidimensional assessment of neonatal pain \cite{zamzmi2017automated}. Similarly, several studies have been conducted to detect pain in adults \cite{walter2013biovid, lu2023transformer, pouromran2021exploration, gupta2025PainXtract}. However, to the best of our knowledge, very few works have explored the rPPG technique for detecting pain in adults \cite{yang2021non}, and that too has not been applied to neonates either.

Unfortunately, these physiological parameters are traditionally measured using contact-based sensors such as photoplethysmography (PPG), ballistocardiogram (BCG), and electrocardiogram (ECG), which may not be suitable for all scenarios. For instance, these contact-based techniques are unsuitable in scenarios such as monitoring neonates with skin damage and infections or while the neonate is sleeping \cite{gupta2023radiant, gupta2017accurate, sahoo2022deep}. Furthermore, the risk of spreading ailments like COVID-19 through sensor contact highlights the need for non-contact alternatives in pandemic conditions \cite{saikia2023hreadai, lokendra2022and}. To this end, rPPG emerges as a non-contact alternative, leveraging video analysis for physiological parameter estimation without direct skin contact.

\section{Proposed Method}

\begin{figure*}[!h]
    \centering
    \includegraphics[width = 0.8\textwidth]{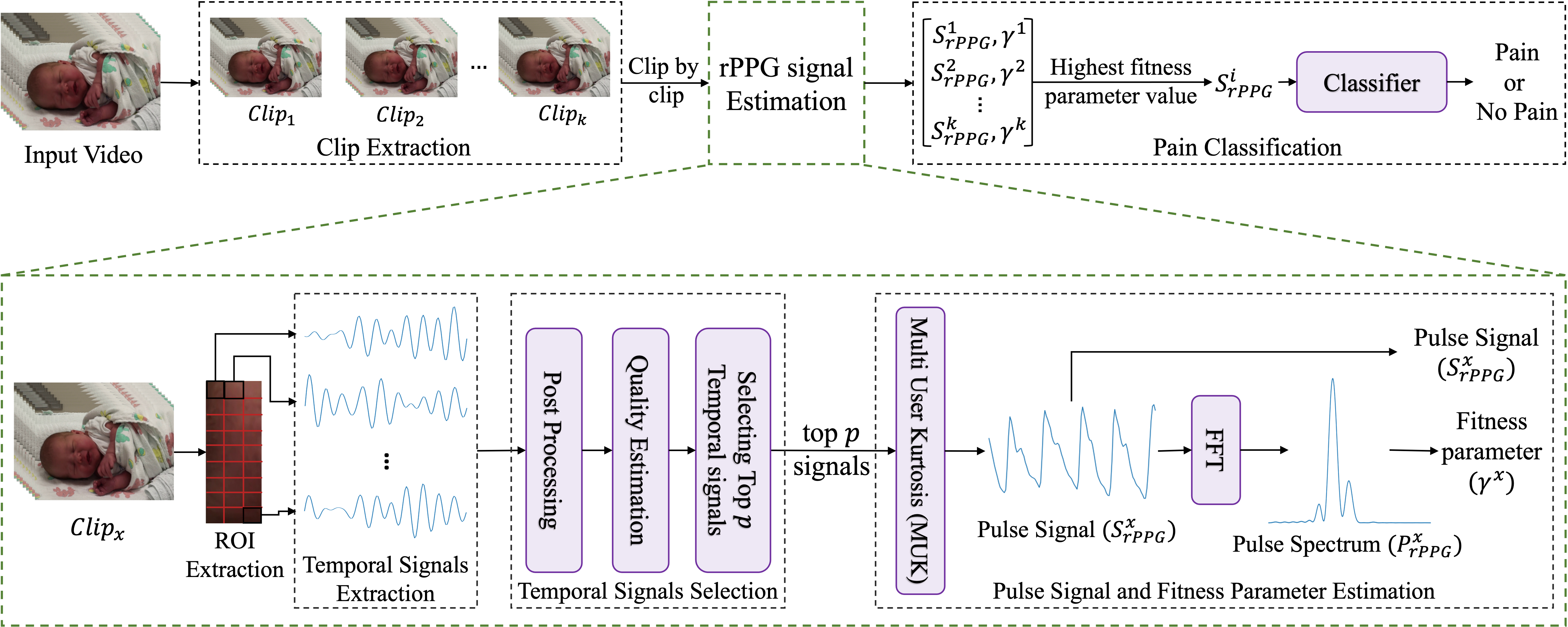}
    \caption{The flow diagram of our proposed method. The input video is divided into multiple clips, the face region is detected and divided into ROIs, and temporal signals with higher quality are selected. The selected signals are used to extract the rPPG signal along with the fitness parameter. Finally, the rPPG signal with the highest fitness parameter is fed to the machine learning classifier for pain and no pain classification.}
    \label{fig:proposed_method}
\end{figure*}

Our proposed method aims to detect neonatal pain utilising rPPG signal extracted from face videos. A flow diagram of the proposed method is shown in Fig \ref{fig:proposed_method}. Initially, the input video is divided into multiple non-overlapping clips. Subsequently, the rPPG signal is extracted from each clip. To this end, the face region is detected, cropped, and divided into multiple ROIs. Temporal signal corresponding to each ROI is then extracted, along with its quality. The quality is defined such that the temporal signals extracted from the ROIs impacted by skin deformations exhibit lower quality due to significant colour variations. Subsequently, the top $p$ signals are selected based on the quality of the temporal signal. The selected temporal signals are then used to extract the rPPG signal along with the fitness parameter. The fitness parameter provides a low value if the clip contains facial deformations. Eventually, several rPPG signals are extracted so that each rPPG signal corresponds to a clip. Some of these signals are obtained from the clips containing noise, exhibiting low values for the fitness parameter. Also, the rPPG signal exhibiting the lowest deformation can provide the most useful information. Therefore, the rPPG signal with the highest fitness parameter is selected and fed to the machine learning (ML) classifier for pain and no pain classification.

\subsection{Clip Extraction}

Few frames in long-duration videos are usually impacted by inevitable noises, motion artefacts, and facial expressions, which can corrupt the rPPG signal \cite{gupta2020mombat}. To address this issue, \cite{gupta2018robust} has divided the input video into multiple clips and chose the clip that is least affected by noise. This way, noisy frames in some clips do not compromise the quality of the rPPG signal extracted from stable clips. Following this, we divided the input video into multiple clips for neonatal pain detection from face videos. In this direction, a non-overlapping window is applied to the video, dividing it into $k$ clips, each consisting of $f$ frames.

\subsection{rPPG Signal Estimation for a Clip}

\subsubsection{ROI extraction} \label{subsec:face_detection_and_roi_extraction}

After dividing the input video into multiple clips, our method estimates the rPPG signal corresponding to each clip. In this direction, our method initially performs neonatal face detection by employing the YOLOv7Face detection model, which is an extension of the YOLOv7 \cite{wang2023yolov7} architecture. It is used for detecting the face and facial landmark points. While the entire facial region of the neonates can be used for temporal signal extraction, noises induced due to eye blinking, lip quivering, and facial expression can impact the quality of the extracted temporal signal \cite{gupta2023radiant, zhao2020remote}. Hence, we select only the forehead region to obtain the rPPG information. Further, the entire forehead region can be considered as a single ROI for the extraction of temporal signals, but different forehead regions exhibit varying degrees of colour variation that can cause significant performance degradation \cite{gupta2017accurate, gupta2023radiant}. Hence, it is recommended in previous studies \cite{gupta2017accurate, gupta2023radiant}, to use multiple smaller ROIs, extract the temporal signals from these smaller ROIs, and finally employ the obtained temporal signals for estimating the pulse signal. Going along the same line, we divide the forehead region into multiple non-overlapping square regions to extract the temporal signals. A square region can be considered as an ROI if all its pixels correspond to the skin, similar to the study \cite{gupta2023radiant}. Thus, we get several ROIs in the clip. Also, the previous works have employed a fixed ROI size \cite{gupta2023radiant, saikia2023hreadai}. However, to cater to the difference in forehead size of each neonate, we employ a dynamic ROI size, which is $m$ times the maximum height and width of the forehead of the neonate. The details of selecting the value of $m$ are provided in section \ref{subsec:implementation_details}. Mathematically, the ROI size, $r_s$ is given by, $r_s = m \times max (height_\text{forehead}, width_\text{forehead})$, where $max$ is the maximum operation while $height_\text{forehead}$ and $width_\text{forehead}$ denote the height and width of the forehead, respectively.

\subsubsection{Temporal Signal Extraction}

The temporal signals are extracted from each ROI in two stages: first, the RGB signals are obtained, followed by the extraction of temporal signals from these obtained RGB signals. The RGB signals for each ROI are obtained by averaging the pixel values of the blue, green, and red colour channels. Mathematically, the signal $\boldsymbol{b}_j$, denotes the blue colour channel signal for the $j^{th}$ ROI is defined as follows:

\begin{equation}
\label{equation:rgbsignals}
\boldsymbol{b}_{j} =\;\left(\frac{\sum_{p} b_{j,1}^{p}}{n_{j,1}}, \frac{\sum_{p} b_{j,2}^{p}}{n_{j,2}}, \ldots, \frac{\sum_{p} b_{j,f}^{p}}{n_{j,f}}\right)
\end{equation}
where $b_{j, i}^{p}$ denotes the $p^{th}$ pixel blue channel intensity for the $j^{th}$ ROI within the $i^{th}$ frame. Here, $n_{j, i}$ indicates the number of pixels in the $j^{th}$ ROI of the $i^{th}$ frame, and $f$ denotes the total number of frames in a video clip. Similarly, the green $\boldsymbol{g}_j$ and red channel $\boldsymbol{r}_j$ are computed. Also, the RGB signal for the $j^{th}$ ROI is given by $(\boldsymbol{r}_{j}, \boldsymbol{g}_{j}, \boldsymbol{b}_{j})$. The temporal signals can be extracted from these RGB signals using either individual colour channel signals from the RGB signals or a weighted combination of two or more colour channels. In our study, we explored individual colour channels and a weighted combination of two or more colour channels as temporal signals in ection \ref{sec:results} and observed that the performance of the blue colour channel temporal signals outperformed others.

\subsubsection{Temporal Signal Selection}\label{subsec:temporal_signal_selection}
The obtained temporal signals ($\boldsymbol{t}_j$) can contain noise due to improper illumination, motion artefacts and facial deformation \cite{gupta2023radiant, saikia2023hreadai, lokendra2022and}. To alleviate the noise, we employ bandpass filtering on the temporal signals with frequencies ranging from 1.67 Hz to 2.91 Hz because neonates' specified HR frequency range corresponds from 100 to 175 beats per minute (BPM). Kindly note that the typical HR range of healthy neonates is highly debated \cite{hutchon2016normal}. Hence, we take a broad range of the HR frequencies based on multiple studies \cite{hutchon2016normal, weissman2012heart, fleming2011normal, nationwidechildrensPhysicalExam}. Mathematically, $ \tilde{\boldsymbol{t}}_j=\psi_{bp}\left(\boldsymbol{t}_j\right)$, where $\psi_{b p}$ denotes the Butterworth bandpass filter, which suppresses the frequencies outside the given frequency range from the signals. 

Subsequently, we select some rPPG signal from the entire set (\(\tilde{t}_1, \tilde{t}_2, \ldots, \tilde{t}_J\)), that are least affected by noise. Our selection process accounts for the susceptibility of temporal signals to noise, especially those influenced by neonatal facial expressions. To this end, we define the quality of each temporal signal such that the temporal signals obtained from ROIs impacted by skin deformation, such as forehead creases, exhibit lower quality. Subsequently, we select those $p$ temporal signals that are least affected by skin deformations, that is, the temporal signals containing higher quality. Hence, we define the quality by leveraging the intuition that the regions with skin deformation will exhibit significant skin colour variations, causing a considerable standard deviation in the amplitude of extracted temporal signals. That is, a temporal signal with a lower standard deviation will provide better rPPG information. Mathematically, $Q_j$ denoting the quality of the $j^{th}$ temporal signal, $\tilde{\boldsymbol{t}}_j$ is calculated by $ Q_j=\frac{1}{\sigma\left(\tilde{\boldsymbol{t}}_j\right)}$, where $\sigma$ denotes the standard deviation operator.

\subsubsection{Pulse Signal and Fitness Parameter Estimation}
The selected top $p$ temporal signals are processed through the Multiuser Kurtosis algorithm \cite{papadias2000multi}, which performs blind source separation on the signals to separate the pulse component and noise. Further, it identifies and extracts the most periodic component, representing the rPPG signal denoted as $S_{rPPG}^{x}$, where $x$ denotes the $x^{th}$ clip of the video. Moreover, we compute the fitness parameter $\gamma^{x}$ corresponding to $\boldsymbol{\mathit{S}}_{rPPG}^{x}$. A high fitness value signifies that the corresponding rPPG signal contains low noise and vice versa \cite{liu2024arobust}. Thus, we define the fitness value by leveraging the intuition that ideally, the amplitude for frequencies other than the HR frequency should be zero, with the highest amplitude value at the HR frequency. However, achieving this ideal scenario is not feasible due to inevitable noise, leading to nonzero amplitude values for frequencies other than HR \cite{saikia2023hreadai}. Hence, we first compute the pulse spectrum, $\boldsymbol{\mathit{P}}_{rPPG}^{x}$ of rPPG signal by applying Fast Fourier Transform (FFT) on $\boldsymbol{\mathit{S}}_{rPPG}^{x}$. Subsequently, the fitness parameter, $\gamma^{x}$ is computed as the SNR ratio of the pulse spectrum, $\boldsymbol{\mathit{P}}_{rPPG}^{x}$.
\begin{equation}
    \gamma^{x} =\frac{\sum_{\nu=m-\alpha}^{m+\alpha} \boldsymbol{\mathit{P}}_{rPPG}^{x}[\nu]}{\sum_{\nu=1.67}^{2.91} \boldsymbol{\mathit{P}}_{rPPG}^{x}[\nu]-\sum_{\nu=m-\alpha}^{m+\alpha} \boldsymbol{\mathit{P}}_{rPPG}^{x}[\nu]} 
\end{equation}
where $m$ denotes the frequency with the highest amplitude in $\boldsymbol{\mathit{P}}_{rPPG}^{x}$, while $\alpha$ denotes the size of the adjacent frequency window. We have selected the frequency range from 1.67 Hz to 2.91 Hz, as discussed above in section \ref{subsec:temporal_signal_selection}.

\subsection{Pain Classification}
Kindly recall that we have divided the input video into $k$ clips. The rPPG signal is extracted, and its fitness parameter is computed for each of the $k$ clips of the video. Among these clips, some clips contain frames affected by unavoidable noise, resulting in lower-quality rPPG signals. Therefore, the value of the fitness parameter tends to be lower for rPPG signals extracted from the clips affected by noise. Hence, from the obtained rPPG signals ($S_{rPPG}^{1}$, $S_{rPPG}^{2}$, \ldots, $S_{rPPG}^{k}$) and their corresponding fitness values ($\gamma^1$, $\gamma^2$, \ldots, $\gamma^k$), we select the rPPG signal ($S_{rPPG}^{i}$) with the highest fitness parameter value, as it is least affected by noise. Mathematically, $i = \mathrm{argmax} \{\gamma^1, \gamma^2, \ldots, \gamma^k\}$. Further, the selected signal ($S_{rPPG}^{i}$) is fed into the classifier, which analyses the signal and provides a binary prediction indicating either pain or no pain. The details of the classifiers are provided in the section \ref{subsec:implementation_details}.
 \section{Experimental Settings}

In this section, we discuss the description of the dataset, implementation details, and the metrics that we have used for the performance evaluation.

\subsection{Dataset Description}\label{sec:dataset}

The iCOPEvid (infant Classification Of Pain Expressions videos) dataset \cite{brahnam2023neonatal} contains 234 videos from 49 neonates, including 23 girls and 26 boys. The neonates in the dataset have an age range of 34 to 70 hours, approximately 1 to 3 days old. These videos were captured in seven distinct settings, wherein four were captured by providing stimuli such as a heel lance puncture, friction on the outer lateral part of the heel, moving the neonate from one crib to another, and an air stimulus to induce an eye squeeze. Whereas the remaining three were captured in between the stimuli, where the neonate was in a resting state. Every video in the dataset is 20 seconds long and labelled as either ``pain" or ``no pain". The dataset contains 49 videos labelled as pain, with one video per infant, and 185 videos labelled as no pain, with multiple videos per infant. However, the publicly available subset of this dataset contains 126 videos labelled as no pain and 49 videos labelled as pain, totalling 175 videos. The dataset is partitioned into training, validation, and testing sets based on the subjects. We have ensured that there is no overlap of subjects between the training, validation, and testing sets to avoid data leakage \cite{saeb2017need}. The training set comprises 158 samples, with 114 labelled as ``no pain" and 44 as ``pain." Meanwhile, both the validation and testing sets consist of 17 samples each, with 12 labelled as ``no pain" and 5 as ``pain." Kindly note that the samples for the validation set were randomly selected.

\subsection{Fusion with Audio modality}\label{sec:fusion_audio}

In our proposed method, the physiological features are computed based on visual features. We validate the efficacy of our proposed method by comparing it to the audio modality for neonatal pain assessment. Further, we combine both physiological and audio features to create a multimodal pain assessment technique and assess its performance. The implementation and results are discussed below.

The audio clip is extracted from the video, and the extracted audio clip is divided into two non-overlapping subclips of $10 s$. The obtained audio sub-clips are processed with a Hanning window with a frame length of $25 ms$, frameshift of $10 ms$, and 128 triangular mel-frequency bins to extract the audio features such as spectrogram, filterbank energy features, and MFCC. These extracted spectral features have been widely employed in various tasks, including emotion detection \cite{wang2022multi, kurpukdee2017study}, and dialect recognition \cite{shon2018convolutional}. The extracted spectral features for each clip are fed to the Swin Transformer \cite{liu2022swin} network based on Vision Transformer (ViT) \cite{dosovitskiy2020image} for pain detection. The probabilities obtained for both the classes corresponding to each subclip are averaged to obtain the final prediction. We have utilised a variant of the base model of the Swin Transformer, which has been pre-trained on the ImageNet-22K dataset \cite{deng2009imagenet} and further fine-tuned on the ImageNet-1K dataset. The model variant was obtained from the Hugging Face repository \cite{wolf2020transformers}, a widely recognised hub for state-of-the-art ML models and resources. The performance is evaluated for all three spectral features, such as spectrogram, filterbank energy features, and MFCC. 

\subsection{Implementation Details} \label{subsec:implementation_details}

The input video is partitioned into $k = 5$ video clips, each containing $f = 120$ frames. These partitioned clips are then passed as input to the YOLOv7Face detection model to detect the face region of the neonate and extract the desired facial region, the forehead. The YOLOv7Face detection model is pre-trained on the Widerface dataset \cite{yang2016wider} and further fine-tuned on the iCOPEvid \cite{brahnam2023neonatal} dataset to increase the accuracy of neonatal face detection, similar to the \cite{grooby2023neonatal}. However, in certain frames, the YOLOv7Face detection model encounters challenges. For instance, in some frames, it fails to detect the neonate's face. Moreover, the YOLOv7Face detection model provides false faces in some of the frames. To mitigate these, we implemented a criterion based on the size of the detected image. If the height and width of a detected image fall below the threshold of $100$ pixels, we consider it as a potentially irrelevant image. In such scenarios, we utilised the coordinates of the previous frame in which the face is detected to compute the forehead region in the current frame. This approach enhances the robustness of our neonatal face detection system, ensuring its applicability in diverse and challenging video scenarios. The obtained forehead region is then divided into multiple ROIs, where the size of the ROI is dynamically selected based on the hyperparameter $m$ and the dimensions of the forehead of the neonate (please refer to section \ref{subsec:face_detection_and_roi_extraction}). We empirically found the best value of $m$ to be $0.05$. The colour channel signals red, green, and blue corresponding to each ROI are obtained. 

The effectiveness of rPPG-based methods is highly dependent on the selection of colour channels as temporal signals for the estimation of the rPPG signal. We have used the individual colour channel signals (red, blue, green) and their weighted linear combination (denoted by $wRGB$) as temporal signals and evaluated their efficacy for the proposed method. For ease of understanding, we denote the settings red, green, blue and wRGB colour channel signals, used as temporal signals as $S_{red}$, $S_{green}$, $S_{blue}$ and $S_{wRGB}$ respectively. For each setting, the rPPG signals obtained corresponding to the training subset of the dataset are used to train a set of ML classifiers for pain and no pain classification. The classifiers are then evaluated on the validation subset, and the classifier with the highest accuracy is chosen. To this end, we have employed the Tree-based Pipeline Optimization Tool (TPOT) \cite{le2020scaling, olson2016evaluation}. TPOT is an automated machine learning (AutoML) tool that uses genetic algorithms to optimize ML pipelines. It is built on top of scikit-learn \cite{scikit-learn}, a popular machine-learning library in Python. TPOT explores various preprocessing steps, feature selectors, classifiers, and hyperparameters to find the optimal pipeline. For each setting of temporal signals, TPOT starts by generating a population of random classifiers, and each classifier is evaluated using $k$-fold cross-validation. This involves splitting the training subset into $k$ folds, training the classifier on $k-1$ folds, and assessing its performance on the remaining fold. Once the training process is complete, TPOT selects the optimal classifier based on a scoring metric. We have used the default parameter values of TPOT, with $k=5$ and the scoring metric as accuracy, to determine the optimal classifier. The chosen classifier is used for inference on the test subset and to compute the performance metrics. Experimentally, we find that the Gaussian Naive Bayes classifier was the most suitable for the setting $S_{red}$, the Random Forest Classifier was the best fit for $S_{blue}$, and the Decision Tree Classifier proved to be the most effective for both the $S_{green}$ and $S_{wRGB}$ settings.

\subsection{Evaluation Metrics}

We employ the standard classification metrics: Accuracy, Precision, Recall, and F1-Score, to assess the performance of the classification. Please note that the metrics Precision, Recall, and F1-Score are calculated using two different averaging techniques: macro and weighted averaging. 

\section{Performance Analysis}\label{sec:results}

\subsection{Overall Performance Across Colour Channels}

We assess the performance of our proposed method on the iCOPEvid dataset \cite{brahnam2023neonatal} for binary classification. The results obtained for the settings $S_{red}$, $S_{green}$, $S_{blue}$ and $S_{wRGB}$ (described in section \ref{subsec:implementation_details}) are reported in Tables \ref{tab:RED}, \ref{tab:GREEN}, \ref{tab:BLUE}, and \ref{tab:RGB}, respectively. Kindly note that we also report the performance of every colour channel as temporal signals for every setting. For instance, the classifier obtained for the setting $S_{red}$, that is, the Gaussian Naive Bayes classifier, is trained on the training subset of every colour and tested on the corresponding testing subset. It can be observed from the Tables \ref{tab:RED}, \ref{tab:GREEN}, \ref{tab:BLUE}, and \ref{tab:RGB} that the performance of rPPG signals extracted from the blue colour channel consistently outperforms the performance of rPPG signals extracted from other colour channels across all the settings. As an instance, for the setting $S_{red}$, the macro and weighted averaged F1-score are 83.65\% and 87.22\%, respectively, with an accuracy of 88.24\% (kindly refer to Table \ref{tab:RED}). The second-best weighted F1-score and accuracy are obtained for the green colour channel, with values being 65.92\% and 70.59\%, respectively. On the other hand, the second-best macro F1-score of 57.50\% is obtained by the red colour channel. Similarly, for the settings, $S_{green}$ (Table \ref{tab:GREEN}) and $S_{wRGB}$ (Table \ref{tab:RGB}), the blue colour channel again outperforms others in terms of macro F1-Score and weighted F1-Score, with the values 79.84\% and 82.77\%, respectively. For accuracy metric, the blue and $wRGB$ colour channels tie for the best score of 82.35\%. Kindly note that among all the settings, the best performance is obtained in the setting $S_{blue}$ and the blue colour channel (refer to Table \ref{tab:BLUE}). For this combination, the accuracy, macro F1-Score and weighted F1-Score are 94.12\%, 92.44\% and  93.91\%, respectively.

\begin{table}[!htbp]
\centering
\caption{Results on every colour channel for the setting $S_{red}$ }
\label{tab:RED}
\begin{tabular}{@{}lccccccc@{}}
\toprule
               & \multicolumn{1}{l}{} & \multicolumn{3}{c}{\textbf{Macro Average}}       & \multicolumn{3}{c}{\textbf{Weighted Average}}    \\ \midrule
\textbf{Channel} &
  \multicolumn{1}{l}{\textbf{Accuracy}} &
  \multicolumn{1}{l}{\textbf{Precision}} &
  \multicolumn{1}{l}{\textbf{Recall}} &
  \multicolumn{1}{l}{\textbf{F1-Score}} &
  \multicolumn{1}{l}{\textbf{Precision}} &
  \multicolumn{1}{l}{\textbf{Recall}} &
  \multicolumn{1}{l}{\textbf{F1-Score}} \\
  \midrule
\textbf{Red}   & 64.71                & 57.50          & 57.50          & 57.50          & 64.71          & 64.71          & 64.71          \\
\textbf{Green} & 70.59                & 61.67          & 55.83          & 55.03          & 66.47          & 70.59          & 65.92          \\
\textbf{Blue}  & \textbf{88.24}       & \textbf{92.86} & \textbf{80.00} & \textbf{83.65} & \textbf{89.92} & \textbf{88.24} & \textbf{87.22} \\
\textbf{wRGB}   & 64.71                & 52.38          & 51.67          & 50.96          & 60.22          & 64.71          & 61.65          \\ \bottomrule
\end{tabular}%
\end{table}

\begin{table}[!htbp]
\centering
\caption{Results on every colour channel for the setting $S_{green}$ }
\label{tab:GREEN}
\begin{tabular}{@{}lccccccc@{}}
\toprule
               & \multicolumn{1}{l}{} & \multicolumn{3}{c}{\textbf{Macro Average}}       & \multicolumn{3}{c}{\textbf{Weighted Average}}    \\ \midrule
\textbf{Channel} &
  \multicolumn{1}{l}{\textbf{Accuracy}} &
  \multicolumn{1}{l}{\textbf{Precision}} &
  \multicolumn{1}{l}{\textbf{Recall}} &
  \multicolumn{1}{l}{\textbf{F1-Score}} &
  \multicolumn{1}{l}{\textbf{Precision}} &
  \multicolumn{1}{l}{\textbf{Recall}} &
  \multicolumn{1}{l}{\textbf{F1-Score}} \\
  \midrule
\textbf{Red}   & 64.71                & 61.43          & 63.33          & 61.36          & 69.08          & 64.71          & 66.04          \\
\textbf{Green} & 64.71                & 57.50          & 57.50          & 57.50          & 64.71          & 64.71          & 64.71          \\
\textbf{Blue}  & \textbf{82.35}       & 78.79          & \textbf{81.67} & \textbf{79.84} & \textbf{83.78} & \textbf{82.35} & \textbf{82.77} \\
\textbf{wRGB}   & \textbf{82.35}       & \textbf{79.81} & 75.83          & 77.33          & 81.79          & \textbf{82.35} & 81.73          \\ \bottomrule
\end{tabular}
\end{table}

\begin{table}[!htbp]
\centering

\caption{Results on every colour channel for the setting $S_{blue}$ }
\label{tab:BLUE}
\begin{tabular}{@{}lccccccc@{}}
\toprule
               & \multicolumn{1}{l}{\textbf{}} & \multicolumn{3}{c}{\textbf{Macro Average}}       & \multicolumn{3}{c}{\textbf{Weighted Average}}    \\ \midrule
\textbf{Channel} &
  \multicolumn{1}{l}{\textbf{Accuracy}} &
  \multicolumn{1}{l}{\textbf{Precision}} &
  \multicolumn{1}{l}{\textbf{Recall}} &
  \multicolumn{1}{l}{\textbf{F1-Score}} &
  \multicolumn{1}{l}{\textbf{Precision}} &
  \multicolumn{1}{l}{\textbf{Recall}} &
  \multicolumn{1}{l}{\textbf{F1-Score}} \\
  \midrule
\textbf{Red}   & 70.59                         & 69.44          & 73.33          & 68.86          & 77.45          & 70.59          & 71.88          \\
\textbf{Green} & 70.59                         & 65.91          & 67.50          & 66.40          & 72.46          & 70.59          & 71.29          \\
\textbf{Blue}  & \textbf{94.12}                & \textbf{96.15} & \textbf{90.00} & \textbf{92.44} & \textbf{94.57} & \textbf{94.12} & \textbf{93.91} \\
\textbf{wRGB}   & 58.82                         & 53.03          & 53.33          & 52.96          & 61.14          & 58.82          & 59.80          \\ \bottomrule
\end{tabular}%
\end{table}

\begin{table}[!h]
\centering
\caption{Results on every colour channel for the setting $S_{wRGB}$ }
\label{tab:RGB}
\begin{tabular}{@{}lccccccc@{}}
\toprule
               & \multicolumn{1}{l}{} & \multicolumn{3}{c}{\textbf{Macro Average}}       & \multicolumn{3}{c}{\textbf{Weighted Average}}    \\ \midrule
\textbf{Channel} &
  \multicolumn{1}{l}{\textbf{Accuracy}} &
  \multicolumn{1}{l}{\textbf{Precision}} &
  \multicolumn{1}{l}{\textbf{Recall}} &
  \multicolumn{1}{l}{\textbf{F1-Score}} &
  \multicolumn{1}{l}{\textbf{Precision}} &
  \multicolumn{1}{l}{\textbf{Recall}} &
  \multicolumn{1}{l}{\textbf{F1-Score}} \\
  \midrule
\textbf{Red}   & 64.71                & 61.43          & 63.33          & 61.36          & 69.08          & 64.71          & 66.04          \\
\textbf{Green} & 64.71                & 57.50          & 57.50          & 57.50          & 64.71          & 64.71          & 64.71          \\
\textbf{Blue}  & \textbf{82.35}       & 78.79          & \textbf{81.67} & \textbf{79.84} & \textbf{83.78} & \textbf{82.35} & \textbf{82.77} \\
\textbf{wRGB}   & \textbf{82.35}       & \textbf{79.81} & 75.83          & 77.33          & 81.79          & \textbf{82.35} & 81.73          \\ \bottomrule
\end{tabular}%
\end{table}

In the literature, the green colour channel has been considered to contain the strongest photoplethysmographic information \cite{gupta2020mombat, saikia2023hreadai}. It provides the most satisfactory signal quality and shows high resistance to motion and lighting artefacts \cite{shchelkanova2021blue}. Although deoxyhemoglobin and oxyhemoglobin absorb most strongly in the blue spectrum at 420 $\mu m$ and 410 $\mu m$, respectively \cite{gajinov2013optical}, blue wavelengths are often disregarded in PPG applications due to their inadequate penetration depth \cite{shchelkanova2021blue}. Furthermore, the blue channel rPPG signal provides comparable or potentially superior results on skin with thinner epidermis \cite{shchelkanova2021blue}. This aspect is particularly relevant for the extraction of rPPG signals in neonates, as their skin composition differs significantly from that of adults. The epidermis of neonates is approximately 20\% thinner than that of adults. Similarly, their stratum corneum, which is the outermost layer of the epidermis, is approximately 30\% thinner. Likewise, the neonate's stratum basale, the deepest layer of the epidermis, is underdeveloped and thinner compared to adults \cite{rahma2022skin}. Our findings support these facts, and we get our best performance when we use the blue colour channel temporal signals to extract rPPG signals.

To strictly evaluate the robustness of our method, we also performed a Leave-One-Subject-Out (LOSO) cross-validation. This protocol ensures that every subject is used for testing exactly once, with strictly no data leakage between training and testing. For this experiment, we utilized the subset of 34 subjects out of the 49 subjects that contained samples for both target labels, with 126 videos for no pain and 34 videos for pain, resulting in a total of 160 videos. We obtained a mean accuracy of 91.96\%, with a 95\% confidence interval (CI) of $[88.73, 95.19]$, which confirms the statistical stability of our results. Please note that these experiments are performed in the setting $S_{blue}$.

\subsection{Effect of Temporal Signal Selection}

\begin{table}[!htb]
\centering
\caption{Results on different values of top $p$ temporal signals}
\label{tab:different_p_values}
\begin{tabular}{@{}cccccccc@{}}
\toprule
                & \multicolumn{1}{l}{} & \multicolumn{3}{c}{\textbf{Macro Average}}       & \multicolumn{3}{c}{\textbf{Weighted Average}}    \\ \midrule
\textbf{\begin{tabular}[c]{@{}c@{}}Number \\ of top $p$ signals \end{tabular}} &
  \multicolumn{1}{l}{\textbf{Accuracy}} &
  \multicolumn{1}{l}{\textbf{Precision}} &
  \multicolumn{1}{l}{\textbf{Recall}} &
  \multicolumn{1}{l}{\textbf{F1-Score}} &
  \multicolumn{1}{l}{\textbf{Precision}} &
  \multicolumn{1}{l}{\textbf{Recall}} &
  \multicolumn{1}{l}{\textbf{F1-Score}} \\
  \midrule
\textbf{top 10} & 88.24                & 85.83          & 85.83          & 85.83          & 88.24          & 88.24          & 88.24          \\
\textbf{top 15} & \textbf{94.12}       & \textbf{96.15} & \textbf{90.00} & \textbf{92.44} & \textbf{94.57} & \textbf{94.12} & \textbf{93.91} \\
\textbf{top 20} & 70.59                & 61.67          & 55.83          & 55.03          & 66.47          & 70.59          & 65.92          \\
\textbf{all}    & 58.82                & 47.12          & 47.50          & 47.11          & 56.22          & 58.82          & 57.36          \\ \bottomrule
\end{tabular}%
\end{table}

We evaluate the performance of our proposed approach across different numbers of top $p$ signals that are least affected by skin deformations. We perform our experiments across four configurations. In the first three, the value of $p$ is set to 10, 15, and 20, respectively. Whereas, in the fourth, we select all the obtained signals for further processing. Note that these experiments are performed in the setting $S_{blue}$, the results for which are reported in Table \ref{tab:different_p_values}. The best performance is obtained for $p=15$. As the number of signals increases beyond 15, there is a noticeable decline in performance. The accuracy drops to 70.59\% for the top 20 and further down to 58.82\% when all signals are considered. The results indicate that including more signals, especially those more affected by skin deformations, introduces noise or irrelevant information, thereby reducing the model's ability to make accurate predictions. On the other hand, when the number of temporal signals is reduced, performance decreases due to the loss of pertinent temporal signals.

\subsection{Effect of Number of Frames}

To analyse the temporal dynamics of rPPG estimation, we evaluated how varying the number of frames per video clip influences the performance of our method. The corresponding results are summarised in Table~\ref{tab:varying_number_of_frames}. We perform this study in five settings, dividing the video into $k=3, 4, 5, 6$ and $7$ clips to obtain the frame count per video clip as 200, 150, 120, 100, and 85, respectively. Please note that for $k=7$, we discard the five remaining frames from the end of each video. We perform this ablation in $S_{blue}$ setting, wherein the Random Forest Classifier is employed, and blue colour channel signals are used to extract the rPPG signal. In addition to these five settings, we have also employed the entire video within a single clip to extract the rPPG signal. It can be observed from the table that the best performance is obtained when the video clip is divided into 120 frames. However, as the number of frames per video clip increases, performance tends to degrade. This is due to the inclusion of frames affected by noise, which can corrupt the rPPG signal \cite{gupta2020mombat}. Similarly, when the frame count is reduced from 120 frames, the performance of the method begins to deteriorate. The accuracy drops from 94.12\% to 82.35\% and 70.59\%, respectively, when we use 100 and 85 frames instead of 120 frames. This behaviour can be attributed to the insufficient rPPG information in video clips with fewer frames \cite{lokendra2022and, saikia2023hreadai}.

\begin{table}[!ht]
\centering
\caption{Results on varying number of frames in a video clip}
\label{tab:varying_number_of_frames}
\begin{tabular}{@{}cccccccc@{}}
\toprule
\multicolumn{1}{l}{} & \multicolumn{1}{l}{} & \multicolumn{3}{c}{\textbf{Macro Average}}       & \multicolumn{3}{c}{\textbf{Weighted Average}}    \\ \midrule
\textbf{\begin{tabular}[c]{@{}c@{}}Number \\ of Frames\end{tabular}} &
  \textbf{Accuracy} &
  \textbf{Precision} &
  \textbf{Recall} &
  \textbf{F1-Score} &
  \textbf{Precision} &
  \textbf{Recall} &
  \textbf{F1-Score} \\
  \midrule
\textbf{85}          & 70.59                & 61.67          & 55.83          & 55.03          & 66.47          & 70.59          & 65.92          \\
\textbf{100}         & 82.35                & 79.81          & 75.83          & 77.33          & 81.79          & 82.35          & 81.73          \\
\textbf{120}         & \textbf{94.12}       & \textbf{96.15} & \textbf{90.00} & \textbf{92.44} & \textbf{94.57} & \textbf{94.12} & \textbf{93.91} \\
\textbf{150}         & 76.47                & 71.67          & 71.67          & 71.67          & 76.47          & 76.47          & 76.47          \\
\textbf{200}         & 76.47                & 72.62          & 65.83          & 67.31          & 75.07          & 76.47          & 74.43          \\
\textbf{all}         & 64.71                & 34.38          & 45.83          & 39.29          & 48.53          & 64.71          & 55.46          \\
\bottomrule
\end{tabular}%
\end{table}

\subsection{Study on Heart Rate Variability Features}

The extracted features from rPPG signals, such as HRV, provide interpretable information and are used for various classification tasks \cite{ding2026detecting}. However, studies have indicated that HRV features extracted from PPG segments shorter than 90 seconds are generally unreliable \cite{mejia2023duration, shaffer2020critical}. Literature defines PPG signals of the length of ~5 min as short-term and signals of the length of less than 5 min as ultra-short-term signals for calculating HRV \cite{shaffer2017overview}. 
Unfortunately, the dataset used in our study comprises video recordings of 20 seconds. Furthermore, we also tried to extract HRV features from our dataset using the widely used library NeuroKit2 toolbox \cite{Makowski2021neurokit}. However, the toolbox failed to compute several HRV parameters, due to the short recording duration. A significant number of HRV features exhibited a high percentage of missing values (NaNs), making them unsuitable for robust classification.

Nevertheless, we recognise the importance of comparing the performance of raw signal-based classification with feature-based methods. Table \ref{tab:HRV} presents a comparison of the classification results obtained using HRV features extracted from a 20-second clip for pain detection and the results from the proposed method. To ensure that only the most relevant and meaningful features were retained, we applied systematic data pre-processing techniques. Specifically, we removed HRV parameters that had more than 70\% missing values, leading to the exclusion of 11 features \texttt{HRV\_SDANN1}, \texttt{HRV\_SDNNI1}, \texttt{HRV\_SDANN2}, \texttt{HRV\_SDNNI2}, \texttt{HRV\_SDANN5}, \texttt{HRV\_SDNNI5}, \texttt{HRV\_ULF}, \texttt{HRV\_VLF}, \texttt{HRV\_LF}, \texttt{HRV\_LFHF}, and \texttt{HRV\_LFn}. Furthermore, we also eliminated 14 low-variance HRV features that lacked meaningful discriminative power, \texttt{HRV\_MCVNN}, \texttt{HRV\_SDRMSSD}, \texttt{HRV\_HF}, \texttt{HRV\_VHF}, \texttt{HRV\_TP}, \texttt{HRV\_PIP}, \texttt{HRV\_IALS}, \texttt{HRV\_PSS}, \texttt{HRV\_C1d}, \texttt{HRV\_C1a}, \texttt{HRV\_Cd}, \texttt{HRV\_Ca}, \texttt{HRV\_MFDFA\_alpha1\_Fluctuation}, and \texttt{HRV\_HFD}. These features either had the same values or extremely little variation across all entries. Once we had removed 25 unsuitable features out of 87 extracted features, we fed the remaining features to \texttt{TPOT}, an AutoML library.

\begin{table}[!htbp]
    \centering
    \caption{Results on every colour channel obtained using filtered HRV features}
    \label{tab:HRV}
    \begin{tabular}{@{}lccccccc@{}}
    \toprule
                   & \multicolumn{1}{l}{} & \multicolumn{3}{c}{\textbf{Macro Average}}       & \multicolumn{3}{c}{\textbf{Weighted Average}}    \\ \midrule
    \textbf{Channel} &
    \multicolumn{1}{l}{\textbf{Accuracy}} &
    \multicolumn{1}{l}{\textbf{Precision}} &
    \multicolumn{1}{l}{\textbf{Recall}} &
    \multicolumn{1}{l}{\textbf{F1-Score}} &
    \multicolumn{1}{l}{\textbf{Precision}} &
    \multicolumn{1}{l}{\textbf{Recall}} &
    \multicolumn{1}{l}{\textbf{F1-Score}} \\
    \midrule
    \textbf{Red}   & 64.71 & 57.50 & 57.50 & 57.50 & 64.71 & 64.71 & 64.71 \\
    \textbf{Green} & 64.71 & 52.38 & 51.67 & 50.96 & 60.22 & 64.71 & 61.65 \\
    \textbf{Blue}  & 70.59 & 61.67 & 55.83 & 55.03 & 66.47 & 70.59 & 65.92 \\
    \textbf{wRGB}  & 58.82 & 47.12 & 47.50 & 47.11 & 56.22 & 58.82 & 57.36 \\ 
    \textbf{Proposed}  & \textbf{94.12}       & \textbf{96.15}         & \textbf{90.00} & \textbf{92.44} & \textbf{94.57} & \textbf{94.12} & \textbf{93.91} \\\bottomrule
    \end{tabular}%
\end{table}

Among the individual colour channels, the blue channel yielded the highest accuracy (70.59\%), whereas the red and green channels showed comparable but lower performances (64.71\%). However, these results are subpar from the results obtained from the raw rPPG signals. This behaviour can be associated with the fact that the HRV estimation using rPPG relies heavily on the accurate detection of inter-beat intervals (IBIs) from the rPPG signal. However, rPPG-based IBIs are often susceptible to missed or false peak detections, which can lead to errors in RR interval calculations and significantly impact key HRV parameters. These results clearly demonstrate that raw rPPG signals retain richer discriminative information compared to derived HRV features, particularly in short-duration recordings.

\subsection{Analysis of Audio Features for Pain Detection}

\begin{table}[!ht]
\centering
\caption{Results on different audio features on Swin Transformer}
\label{tab:diff_audio_features}
\begin{tabular}{@{}lccccccc@{}}
\toprule
                       & \multicolumn{1}{l}{} & \multicolumn{3}{c}{\textbf{Macro Average}}               & \multicolumn{3}{c}{\textbf{Weighted Average}}            \\ \midrule
\textbf{Audio Feature} & \textbf{Accuracy}    & \textbf{Precision} & \textbf{Recall} & \textbf{F1-Score} & \textbf{Precision} & \textbf{Recall} & \textbf{F1-Score} \\
\midrule
Spectrogram & 76.47 & 87.50 & 60.00 & 59.52 & 82.35 & 76.47 & 70.31 \\
Filterbank            & \textbf{88.24}       & \textbf{92.86}     & \textbf{80.00}  & \textbf{83.65}    & \textbf{89.92}     & \textbf{88.24}  & \textbf{87.22}    \\
MFCC        & 82.35 & 90.00 & 70.00 & 73.02 & 85.88 & 82.35 & 79.55 \\ \bottomrule
\end{tabular}%
\end{table}

As discussed in the section \ref{sec:fusion_audio}, we compare the efficacy of the proposed method against the audio modality for pain detection. Further, we also combine both features to create a multimodal pain detection method to assess its efficacy. The results are reported for different audio features for pain detection in Table \ref{tab:diff_audio_features}. We can observe from the table that filterbank energy features outperform other audio features by a significant margin. The spectrogram features attain the lowest performance among the three features, with an accuracy of 76.47\%, a macro average F1-score of 59.52\% and a weighted average F1-score of 70.31\%. The MFCC features perform better than spectrogram features with an accuracy of 82.35\%, a macro average F1-score of 73.02\% and a weighted average F1-score of 79.55\%. The rationale behind this behaviour is that, unlike spectrogram, filterbank energy features replicate how sound is processed by the human ear \cite{zhang2022research, gupta2024radiance}. Filterbank energy features utilise the mel scale, which closely resembles the auditory response of the human ear, making them suitable for tasks like depression detection and emotion recognition. Further, the filterbank energy features undergo a processing step such as Discrete Cosine Transformation to extract the MFCC features \cite{zhang2022research}. This extra step of processing results in the loss of relevant information for tasks like emotion detection \cite{venkataramanan2019emotion}.

\begin{table}[!ht]
\centering
\caption{Results on filterbank energy features for variants of Swin Transformer and fusion of rPPG and Audio}
\label{tab:audio_and_consolidaton}
\begin{tabular}{@{}ccccccccc@{}}
\toprule
\multicolumn{1}{l}{} &
  \multicolumn{1}{l}{} &
  \multicolumn{1}{l}{} &
  \multicolumn{3}{c}{\textbf{Macro Average}} &
  \multicolumn{3}{c}{\textbf{Weighted Average}} \\ \midrule
\textbf{\begin{tabular}[c]{@{}c@{}}Model \\ Variant\end{tabular}} &
  \textbf{\begin{tabular}[c]{@{}c@{}}Additional \\ Pretraining\end{tabular}} &
  \textbf{Accuracy} &
  \textbf{Precision} &
  \textbf{Recall} &
  \textbf{F1-Score} &
  \textbf{Precision} &
  \textbf{Recall} &
  \textbf{F1-Score} \\
  \midrule
Tiny &
  No &
  76.47 &
  72.62 &
  65.83 &
  67.31 &
  75.07 &
  76.47 &
  74.43 \\
Small &
  No &
  82.35 &
  79.81 &
  75.83 &
  77.33 &
  81.79 &
  82.35 &
  81.73 \\
Base &
  No &
  82.35 &
  78.79 &
  81.67 &
  79.84 &
  83.78 &
  82.35 &
  82.77 \\
Base &
  Yes &
  \underline{88.24} &
  \underline{92.86} &
  \underline{80.00} &
  \underline{83.65} &
  \underline{89.92} &
  \underline{88.24} &
  \underline{87.22} \\
  \midrule
\multicolumn{2}{c}{rPPG} &
  \textbf{94.12} &
  \textbf{96.15} &
  90.00 &
  92.44 &
  94.57 &
  \textbf{94.12} &
  93.91 \\
  \midrule
\multicolumn{2}{c}{\begin{tabular}[c]{@{}c@{}}\textbf{Fusion} \\ \textbf{(rPPG + Audio)} \end{tabular}} &
  \textbf{94.12} &
  91.67 &
  \textbf{95.83} &
  \textbf{93.28} &
  \textbf{95.10} &
  \textbf{94.12} &
  \textbf{94.26} \\ \bottomrule
\end{tabular}%
\end{table}

Further, to study the effect of model size and complexity, we have utilized three more variants of Swin Transformer: tiny, small, and base, all trained on the ImageNet-1K dataset \cite{imagenet15russakovsky}. All these models are also obtained from the Hugging Face repository \cite{wolf2020transformers}. The results for filterbank energy features for all four variants of Swin Transformer model are reported in Table \ref{tab:audio_and_consolidaton}. From the table, we can observe that the performance improves when the size and complexity of the model increase. The base model with additional pre-training demonstrated superior performance compared to the other models. The best results are obtained for the base variant with additional pre-training, attaining accuracy, macro F1-Score  and weighted F1-Score of 88.24\%, 83.65\% and 87.22\%, respectively. The best scores obtained for the audio modality are underlined in the table for ease of reference. Further, we also consolidate the information obtained from the rPPG and the audio modalities to study the efficacy of the proposed method. To this end, the log probabilities obtained from the rPPG and audio modalities are concatenated and fed to a Support Vector Classifier (SVC) to obtain the final prediction. The SVC performs binary classification, classifying the input as indicating pain or no pain. We perform the fusion in $S_{blue}$ setting, wherein the Random Forest Classifier is employed, and the rPPG signals are extracted using blue colour channel signals. The results of the fusion are reported in the last row of Table \ref{tab:audio_and_consolidaton}. The fusion of rPPG and audio features results in even higher F1-scores of 93.28\% (macro) and 94.26\% (weighted), surpassing the performance of the individual modalities.

\section{Comparative Analysis}\label{sec13}
We compare our results with existing works in Table \ref{tab:comparitive}. For the methods proposed in \cite{brahnam2023neonatal} and \cite{lu2024video}, we implemented the architectures based on the details provided in their respective papers. For the approaches in \cite{ferreira2025disclosing}, we utilized the publicly available code provided by the authors. Kindly note that we did not use any extra datasets for pre-training the models; however, we followed data augmentation to increase the number of training samples and improve generalization. Further, we employed MediaPipe to extract faces across all experiments to ensure uniformity in preprocessing. As observed in the table, our proposed method significantly outperforms the competing approaches. While the two-stream architecture (TSCN-CSA) \cite{lu2024video} achieves a competitive accuracy of 88.24\%, our method achieves the highest accuracy of 94.12\%. This demonstrates that extracting rPPG signals from the blue channel yields superior discriminative power for neonatal pain detection compared to methods relying solely on facial features.

\begin{table}[!htbp]
\centering
\caption{Performance comparison of the proposed method with existing methods}
\label{tab:comparitive}
\begin{tabular}{@{}lccccccc@{}}
\toprule
               & \multicolumn{1}{l}{} & \multicolumn{3}{c}{\textbf{Macro Average}}       & \multicolumn{3}{c}{\textbf{Weighted Average}}    \\ \midrule
\textbf{Method} &
  \multicolumn{1}{l}{\textbf{Accuracy}} &
  \multicolumn{1}{l}{\textbf{Precision}} &
  \multicolumn{1}{l}{\textbf{Recall}} &
  \multicolumn{1}{l}{\textbf{F1-Score}} &
  \multicolumn{1}{l}{\textbf{Precision}} &
  \multicolumn{1}{l}{\textbf{Recall}} &
  \multicolumn{1}{l}{\textbf{F1-Score}} \\
  \midrule
Baseline \cite{brahnam2023neonatal}  & 76.47& 71.67& 71.67& 71.67& 76.47& 76.47& 76.47\\
VGG-Face \cite{ferreira2025disclosing} & 70.59& 69.44& 73.33& 68.86& 77.45& 70.59& 71.88\\
ViT-B/16 \cite{ferreira2025disclosing} & 70.59& 65.91& 67.50& 66.40& 72.46& 70.59& 71.29\\
TS-ConvNet \cite{lu2024video} & 82.35& 79.81& 75.83& 77.33& 81.79& 82.35& 81.73\\
TSCN-CSA  \cite{lu2024video} & 88.24& 92.86& 80.00& 83.65& 89.92& 88.24& 87.22\\ \midrule
Proposed   & \textbf{94.12}& \textbf{96.15}& \textbf{90.00}& \textbf{92.44}& \textbf{94.57}& \textbf{94.12}& \textbf{93.91}\\ \bottomrule
\end{tabular}%
\end{table}

\section{Conclusion}\label{sec14}
Timely detection of neonatal pain plays a pivotal role in their well-being, as untreated neonatal pain is associated with learning and developmental impairments, brain structure alterations, and slower weight gain. To this end we employed rPPG to non-invasively extract rPPG signals from facial videos and use them to assess neonatal pain. Our study indicated that the rPPG signals extracted using the blue colour channel yield superior performance compared to other colour channels. We utilised a quality parameter to select the temporal signals least affected by skin deformations. Furthermore, we used the SNR ratio as the fitness parameter of the rPPG signal obtained from each clip to select the rPPG signal corresponding to the clip that was least affected by inevitable noises. Our results demonstrated that combining rPPG and audio features provides superior performance compared to individual modalities.

In the future, we plan to estimate physiological signals in a semi-supervised manner to develop a system that can operate reliably in the challenging NICU environment. Further, we aim to incorporate behavioural features such as facial expressions and body movements to improve system accuracy and applicability. We will also analyse results across multiple neonatal datasets and benchmark our method against state-of-the-art approaches. Likewise, we plan to develop models using techniques like feature importance to ensure transparency in decision-making. We also aim to extend this research by collecting simultaneous video and contact-based physiological recordings, such as ECG and PPG. Such multimodal data will enable direct validation of non-contact signals, further establishing the clinical reliability of rPPG-based pain assessment.

\section*{Acknowledgments}
The authors express gratitude to the researchers who have made the iCOPEvid dataset accessible. Anup Kumar Gupta and Trishna Saikai acknowledge the support of the Prime Minister Research Fellowship (PMRF) program of the Government of India. 

\bibliography{refs.bib}

\end{document}